\documentclass[letterpaper, 10 pt, conference]{ieeeconf}
\IEEEoverridecommandlockouts

\usepackage{bm}
\usepackage{amsmath} 
\usepackage{amssymb}  
\usepackage{amsfonts}

\usepackage{subfigure}
\usepackage{graphicx}
\usepackage{hyperref}
\usepackage{cite}

\usepackage{algorithm,algorithmic} 

\usepackage{booktabs}
\usepackage{multirow}
\usepackage{tablefootnote}
\usepackage{threeparttable}

\usepackage{mathtools}
\usepackage{arydshln}
\usepackage{soul, color, xcolor}
\usepackage{pifont}

\newcommand{\xmark}{\ding{55}}%

\definecolor{aureolin}{rgb}{0.99, 0.9, 0.0}
\definecolor{caribbeangreen}{rgb}{0.0, 0.8, 0.6}

\usepackage{xcolor}
\usepackage{colortbl}
\usepackage{caption}
\title{\LARGE \bf
ProtoGuard-guided PROPEL: Class-Aware Prototype Enhancement and Progressive Labeling for Incremental 3D Point Cloud Segmentation
}

\begin{document}

\hbadness=2000000000
\vbadness=2000000000
\hfuzz=100pt

\setlength{\abovedisplayskip}{0pt}
\setlength{\belowdisplayskip}{0pt}
\setlength{\floatsep}{3pt plus 1.0pt minus 1.0pt}
\setlength{\intextsep}{3pt plus 1.0pt minus 1.0pt}
\setlength{\textfloatsep}{3pt plus 1.0pt minus 1.0pt}
\setlength{\parskip}{0pt}

\author{
Haosheng Li\textsuperscript{1},
Yuecong Xu\textsuperscript{2},
Junjie Chen\textsuperscript{1},
Kemi Ding\textsuperscript{1}
\thanks{H. Li, J. Chen and K. Ding are with the Department of Automation and Intelligent Manufacturing (AIM), Southern University of Science and Technology, Shenzhen, China. Email: {\tt\small\{12332662, 12332651\} @mail.sustech.edu.cn, dingkm@sustech.edu.cn.}}
\thanks{Y. Xu is with the Department of Electrical and Computer Engineering, National University of Singapore. Email: 
{\tt\small yc.xu@nus.edu.sg}}}

\maketitle
\begingroup\renewcommand\thefootnote{\textsection}
\endgroup

\begin{abstract}

3D point cloud semantic segmentation technology has been widely used. However, in real-world scenarios, the environment is evolving.
Thus, offline-trained segmentation models may lead to catastrophic forgetting of previously seen classes. Class-incremental learning (CIL) is designed to address the problem of catastrophic forgetting. While point clouds are common, we observe high similarity and unclear boundaries between different classes. Meanwhile, they are known to be imbalanced in class distribution. These lead to issues including misclassification between similar classes and the long-tail problem, which have not been adequately addressed in previous CIL methods. We thus propose ProtoGuard and PROPEL (Progressive Refinement Of PsEudo-Labels). In the base-class training phase, ProtoGuard maintains geometric and semantic prototypes for each class, which are combined into prototype features using an attention mechanism. In the novel-class training phase, PROPEL inherits the base feature extractor and classifier, guiding pseudo-label propagation and updates based on density distribution and semantic similarity. Extensive experiments show that our approach achieves remarkable results on both the S3DIS and ScanNet datasets, improving the mIoU of 3D point cloud segmentation by a maximum of 20.39\% under the 5-step CIL scenario on S3DIS.
\end{abstract}

\section{Introduction}
3D point cloud segmentation methods~\cite{lee2022patchwork++,wu2023generalized} are critical for robotics, enabling object detection, navigation, and environment mapping in dynamic environments. Point clouds present challenges due to their large scale, sparse nature, and high variability in object shapes. However, traditional retraining methods~\cite{klabjan2020neural, wu2020deltagrad} are inefficient and computationally expensive for such dynamic settings, underscoring the need for class-incremental learning (CIL) systems that can adapt efficiently while mitigating these limitations.

\begin{figure}[t]
    \centering
    \includegraphics[width=.85\linewidth]{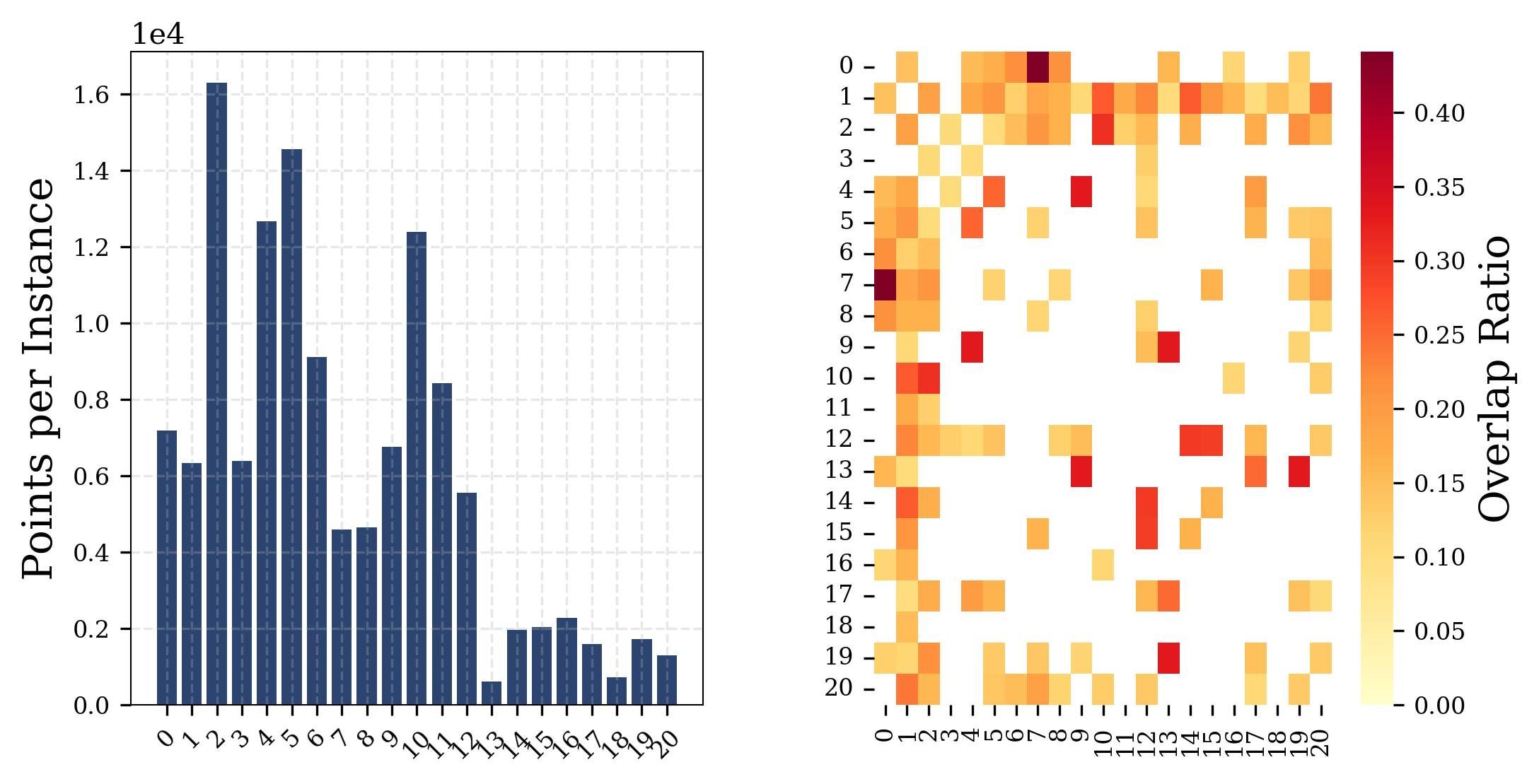}
    \caption{Visualization of the distribution of each class in ScanNet~\cite{dai2017scannet} and the degree of overlap between classes. We define the degree of overlap between two classes as the ratio of points in one class closer than 0.1 meters to the point cloud of another class.}
    \label{Fig: challenge}
\end{figure}

CIL enables models to adapt to new data without catastrophic forgetting. Most existing CIL research focuses on 2D image~\cite{9577808, gao2023ddgr, wen2024class}. However, due to point clouds' irregular and complex structure, research on CIL based on 3D point cloud data is challenging and limited. Thus, most studies focus on more straightforward classification and recognition tasks~\cite{chowdhury2022few, dong2021i3dol, 9981167}. CIL for 3D point cloud segmentation faces two significant challenges, as shown in Fig~\ref{Fig: challenge}. First, segmentation tasks involve more complex scenes, often containing multiple object classes with significant variations in class distribution. Second, objects frequently exhibit overlapping distributions in the 3D point cloud space. CIL for 3D point cloud segmentation is thus more challenging, as catastrophic forgetting is more likely to occur for long-tail classes. Because these classes have sparse representations, they receive infrequent training updates that fail to reinforce their learned features adequately. Meanwhile, misclassification remains common between overlapping classes. 

Our approach addresses these challenges by introducing a two-stage framework that maintains prototypes in base-class training to guide the generation and propagation of pseudo-labels in novel-class training. We aim to mitigate the effects of long-tail distributions and the misclassification between similar base and novel classes by incorporating geometric and semantic information. Specifically, we introduce ProtoGuard for the base-class training phase. It dynamically maintains geometric and semantic prototypes for each class. These prototypes are updated through similarity-based learning, enabling better class discrimination and improving feature representation. Additionally, we propose PROPEL (Progressive Refinement of Pseudo-Labels) for the novel-class training phase. It guides the generation of pseudo-labels using both the model’s predictions along with density and semantic similarity information. By adapting thresholds based on these factors, our method improves the reliability of pseudo-label propagation, allowing the model to focus on long-tail and overlapping regions and refine its predictions over time.

In summary, our contributions are threefold: (a) we propose a class-aware prototype enhancement method, ProtoGuard, which maintains dynamic geometric and semantic prototypes for each class,  improving base class performance during CIL; (b) we further introduce a progressive refinement approach, PROPEL, for pseudo-label generation, leveraging geometric, semantic, and uncertainty information to enhance pseudo-label quality and guide CIL; and (c) we demonstrate the effectiveness of our method in CIL 3D point cloud segmentation tasks, showing significant improvement in handling catastrophic forgetting. On the S3DIS~\cite{armeni20163d} and ScanNet~\cite{dai2017scannet} datasets, our method achieved a maximum mIoU improvement of 11.65\% and 24.28\%, respectively, under different settings of single-step CIL. In the five-step CIL scenario on S3DIS, we achieved a remarkable performance improvement of 20.39\%.

\section{Related Work}
\subsection{Class-Incremental Learning}
Class-incremental learning (CIL) is an essential subfield of continual learning, focusing on addressing the challenge of how models can learn new classes from continuously incoming data while retaining previously acquired knowledge. In contrast to task-incremental learning and domain-incremental learning, CIL is more challenging as it requires models to predict across all known classes during the test phase rather than being restricted to a specific task or domain. We generally categorize the existing CIL methods into the following four categories: (a) \textit{data-level methods} which focus on mitigating catastrophic forgetting by processing data including data replay~\cite{9577808, gao2023ddgr, 9711397} and data regularization~\cite{ahn2019uncertainty, wang2021training}; (b) \textit{parameter-level methods}~\cite{yu2024pi, chaudhry2018riemannian, yang2021cost} which focus on adjusting or constraining the model's parameters via approaches such as parameter regularization; (c) \textit{algorithm-level methods} that aim to enable continuous learning of new knowledge while retaining previous knowledge, more specifically, knowledge distillation~\cite{wen2024class, asadi2023prototype} transfers knowledge from the old model to the new one, while model rectification~\cite{zhi2024dual, he2024gradient, zhou2024revisiting} enhances model adaptability by modifying specific parts; and (d) \textit{template-level methods}~\cite{mcdonnell2024ranpac, shi2023prototype, qiao2024towards} which leverage predefined templates or exemplars to help the model recognize both new and old classes during the prediction phase.

There have been few recent research on CIL in the 3D point cloud domain, with most of the current studies still focusing on point cloud classification~\cite{chowdhury2022few, dong2021i3dol} and recognition~\cite{9981167, knights2022incloud}. Currently, CIL for fine-grained segmentation remains an under-explored task. Yang et al. ~\cite{yang2023geometry} conducted primary research on CIL for 3D point cloud segmentation by focusing solely on the novel-class training phase. In contrast, we further focus on the base-class training phase. By training a better prototype for each class from the base classes, we enhance the model's discriminative power, thereby providing better guidance for novel class training.

\subsection{3D Point Clouds Semantic Segmentation}
Point cloud segmentation assigns semantic labels to individual points in 3D space. PointNet~\cite{qi2017pointnet}, as the pioneering work of point-based networks, treats each point in the point cloud as an independent instance, capturing local information through feature learning for each point and then learning the global feature representation of the entire point cloud by summarizing these local features. Subsequently, Cylinder3D~\cite{zhu2021cylindrical} introduces two key components, cylinder partitioning and asymmetric 3D convolutional networks, which maintain three-dimensional geometric relationships and improve upon traditional methods that project point clouds into 2D space. Meanwhile, KPConv~\cite{thomas2019kpconv} performs convolution operations using core points on point clouds, and by learning the positions of core points in continuous space, it can adapt to various local geometric shapes. A more recent PTv3~\cite{wu2023point} leverages the Transformer's concept, applying attention mechanisms from point cloud processing to segmentation tasks. It scales point cloud data to larger sizes while maintaining high efficiency, achieving good performance in various point cloud processing tasks.

\section{Methodology}

\begin{figure*}[!ht]
    \centering
    \includegraphics[width=.8\linewidth]{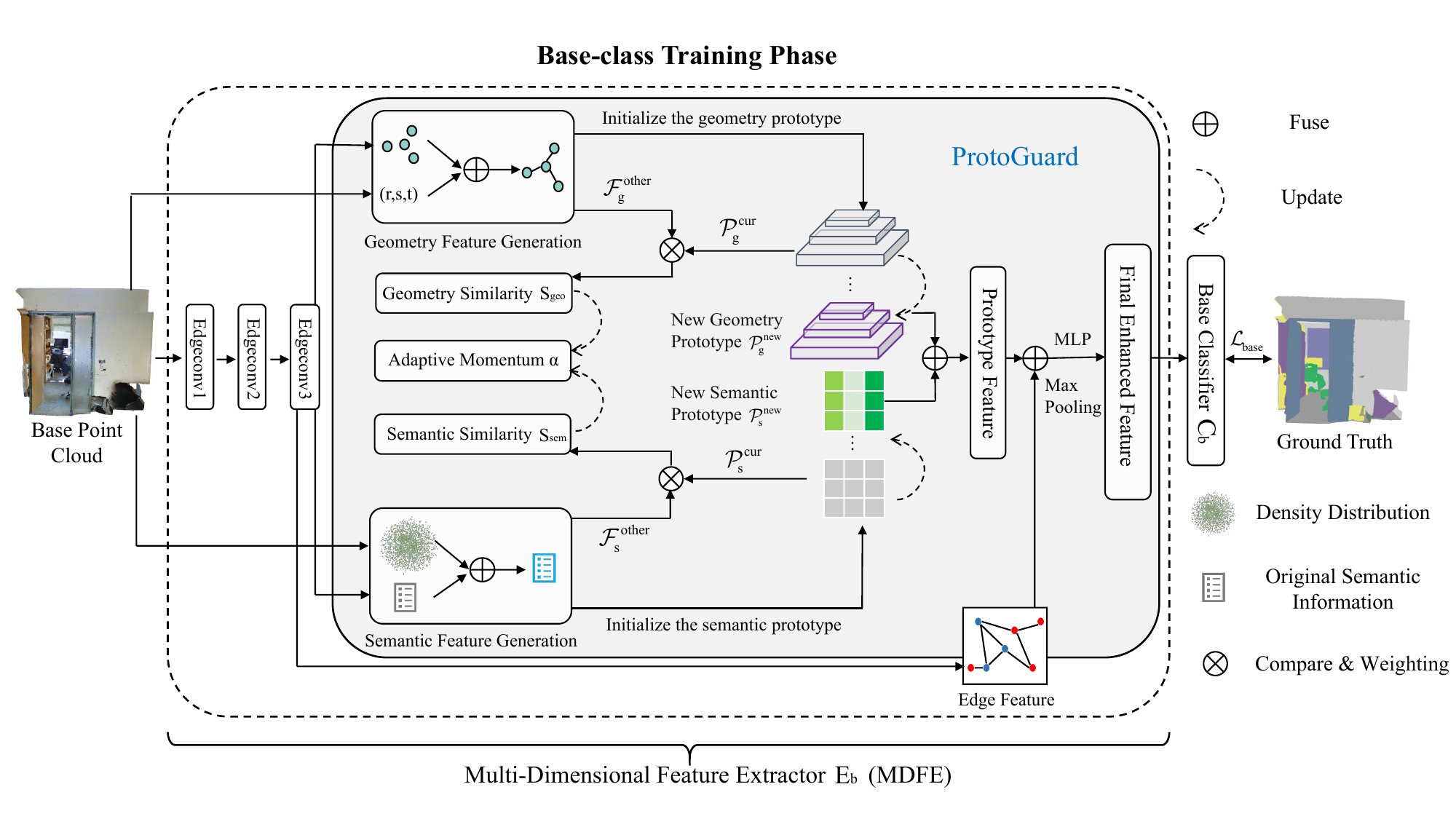}
    \vspace{-6pt}
    \caption{Illustration of ProtoGuard. It is divided into two nearly symmetric geometry-semantic streams, each maintaining a geometric prototype and a semantic prototype for each class. These prototypes improve the model's discriminative ability and when combined with edge features, generate more informative features.}
    \label{Fig: base}
\end{figure*}

\subsection{Problem Setup and Overview}
In this paper, we address the problem of class incremental 3D point cloud segmentation, where the model continuously adapts to new classes without catastrophic forgetting, i.e., forgetting the previously learned classes. A 3D point cloud is represented as a set of points $\mathcal{D}=\left\{d_1, d_2, \ldots, d_N\right\}$, where each point $d_n^i$ is characterized by its spatial coordinates $\left(r_i, s_i, t_i\right)$. For simplicity, we assume that the point cloud is densely sampled. Each point is assigned a label $y_i \in\left\{Y_{\text {base}} \cup Y_{\text {novel}}\right\}$, where $Y_{\text {base}}$ denotes the set of base classes and $Y_{\text {novel}}$ refers to the novel classes introduced incrementally. Since the annotated data for all classes cannot be obtained simultaneously, class-incremental learning (CIL) for 3D point cloud segmentation typically follows a multi-phase approach: a base-class training phase with complete annotations for base classes and a novel-class training phase where new classes emerge with limited annotation resources.

\begin{figure*}[t]
    \centering
    \includegraphics[width=0.8\linewidth]{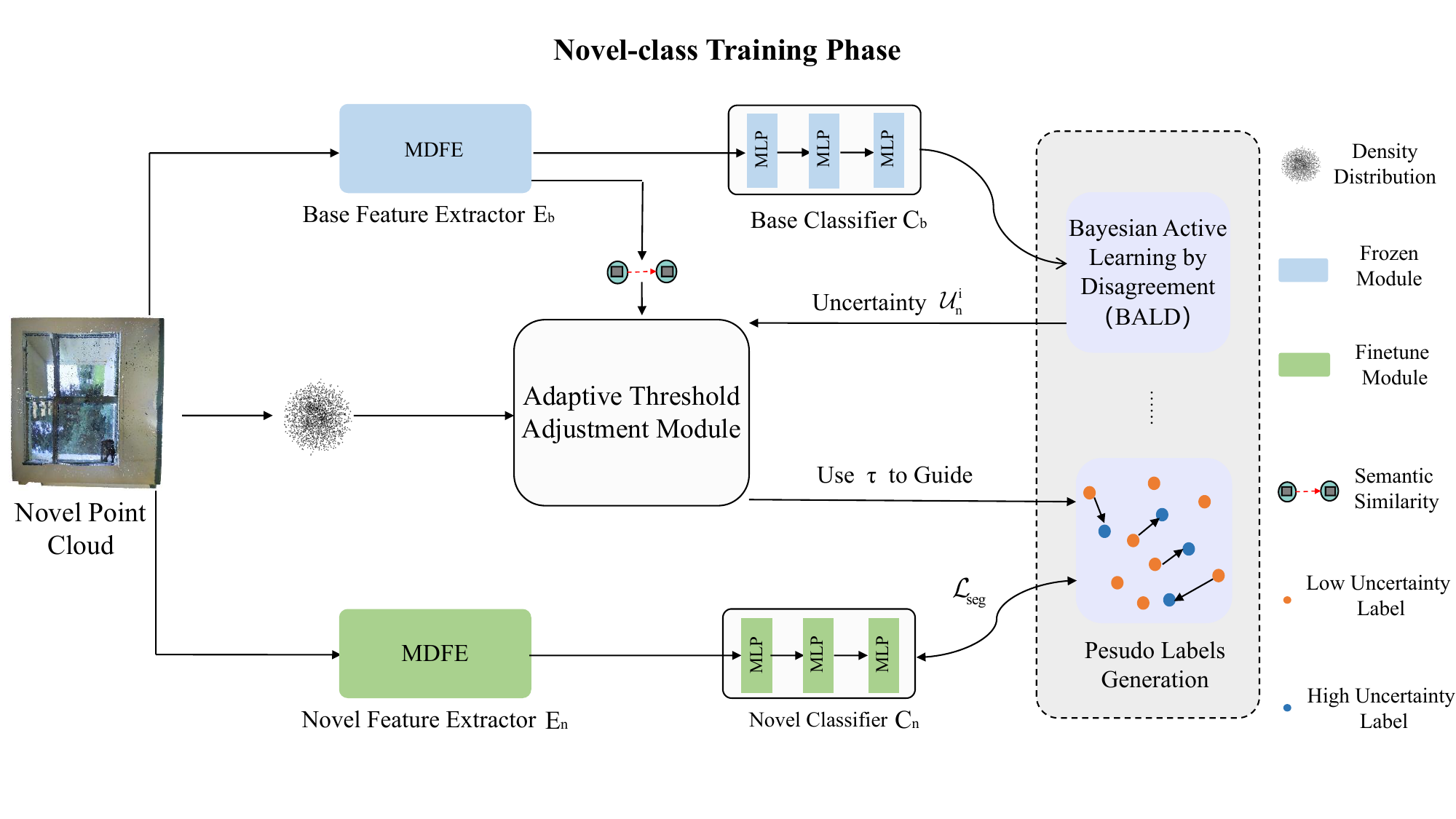}
    \caption{Illustration of PROPEL. The PROPEL framework features a dual-pathway architecture for point cloud CIL. The frozen base model (blue) and trainable novel model (green) process point clouds simultaneously, while our uncertainty estimation module identifies confident (orange) and uncertain (blue) regions. The adaptive threshold dynamically adjusts based on local context, enabling strategic label propagation from confident areas to uncertain regions. }
    \label{Fig: incre}
\end{figure*}

As shown in Fig.~\ref{Fig: base}, in the base-class training phase, we propose the ProtoGuard, which dynamically maintains a geometric prototype $\mathcal{P}_g$ and a semantic prototype $\mathcal{P}_s$ for each class. These prototypes generate prototype features, which are then enhanced by combining them with edge features. This process generates enhanced features, allowing the feature extractor ${E}_b$ and classifier ${C}_b$ to learn more discriminative representations for the base classes. Furthermore, as shown in Fig.~\ref{Fig: incre}, in the novel-class training phase, the model utilizes the feature extractor ${E}_b$ and the classifier ${C}_b$ trained during the base-class training phase, which remains frozen throughout the novel-class training phase. A new novel feature extractor ${E}_n$ and novel classifier ${C}_n$ are fine-tuned on the novel classes. The novel labels' uncertainty $U_i$ is evaluated, and an adaptive threshold adjustment module refines the uncertainty-based guidance for the classifier. High-confidence pseudo labels are then generated and used to train the novel classifier. Next, we dive into the details of each module.

\subsection{ProtoGuard: Geometric and Semantic Prototype Enhancement}
In the base-class training phase, we propose ProtoGuard to improve the discrimination ability of the feature extractor ${E}_b$ and the classifier ${C}_b$. ProtoGuard maintains geometric and semantic prototypes for each class through the following three steps: \textcolor{black}{geometric/semantic prototype initialization, geometric/semantic prototype update, and overall feature fusion.}

First, the geometric and semantic prototypes are initialized from their respective features. As shown in Fig.~\ref{Fig: base}, the geometric feature is combined with the normal vectors and height information. Meanwhile, the semantic feature is computed from the density distribution, derived from the spatial coordinates of the original point cloud, along with the original semantic information. 

Second, we dynamically maintain the prototypes of each class by updating the prototypes since the initial prototype is unstable, which was directly initialized via averaging the individual features of a particular class. To update the prototype, we leverage an adaptive momentum, preventing them from updating too quickly or too slowly. We compute based on geometric and semantic similarities, which are defined as the cosine similarity between the current class prototype and the features of other classes. With the computed similarities, we obtain adaptive momentum, which are learnable adjustment factors. The prototypes are then updated using these momentums. The geometric and semantic features are finally fused through an attention-based mechanism. 

Last, our final feature fusion combines prototype features and edge features, followed by MLP and max pooling. The edge features are obtained by calculating the feature differences between each pair of neighboring points. The network is optimized using the cross-entropy segmentation loss.

\subsection{PROPEL: Progressive Refinement Of PsEudo-Labels}
\textcolor{black}{After the ProtoGuard, which enhances feature representation and classification capabilities for base classes, the challenge in novel-class training lies not only in extending these capabilities to novel classes, but also in preserving the recognition ability of the base classes.} To address these issues, in the novel-class training phase, inspired by~\cite{yang2023geometry}, we introduce PROPEL, which progressively refines pseudo-labels for novel classes while leveraging base-class knowledge.

\textcolor{black}{First, initialize the base/novel feature extractor and base/novel classifier.} Specifically, we inherit the feature extractor and classifier from the base-class training phase to initialize our network structure for the CIL stage. As shown in Fig.~\ref{Fig: incre}, the base feature extractor and classifier parameters are frozen during this phase, preserving their strong extraction and segmentation capabilities for base classes while guiding pseudo-label generation. Simultaneously, we introduce new feature extractor and classifier components that are initialized from the corresponding base components. These initialized components are then fine-tuned to adapt to novel class representations, effectively learning new class information while maintaining base knowledge and leveraging the feature representations already learned during the base-class training phase.

\textcolor{black}{Second, estimate the point-wise uncertainty in novel point clouds.} Here, we apply a spatial sampling approach based on Bayesian Active Learning by Disagreement (BALD)~\cite{houlsby2011bayesian}. BALD effectively captures distribution uncertainty through multiple forward passes with neighborhood configurations. For each point $d_n^i$ in the novel point cloud, uncertainty is computed. It measures the difference between the entropy of the expected prediction and the expected entropy of predictions, quantifying the model's prediction consistency across different spatial contexts. Specifically, we leverage weighted predictions based on spatial proximity. This implementation explicitly models the expectation terms in the theoretical equation by averaging over $T$ different neighborhood configurations while accounting for the spatial structure inherent in point cloud data.

\begin{table*}[t]
\centering
\renewcommand{\arraystretch}{1.2}
\setlength{\tabcolsep}{3pt}
\footnotesize
\caption{The experimental results of 3D point cloud segmentation CIL methods and other necessary comparison methods under split $S^0$ and $S^1$ on the S3DIS dataset, with the best results for CIL methods in this setting highlighted in bold. All results are presented in mIoU (\%).}
\resizebox{0.8\linewidth}{!}{
\begin{minipage}{\linewidth}
\label{Tab: 1}
\begin{tabular*}{\linewidth}{@{\extracolsep{\fill}}c|ccc|ccc|ccc|ccc|ccc|ccc@{}}
\hline
\multirow{3}{*}{Methods} & \multicolumn{6}{c|}{$C_{novel}=5$} & \multicolumn{6}{c|}{$C_{novel}=3$} & \multicolumn{6}{c}{$C_{novel}=1$} \\
\cline{2-19}
& \multicolumn{3}{c|}{$S^0$} & \multicolumn{3}{c|}{$S^1$} & \multicolumn{3}{c|}{$S^0$} & \multicolumn{3}{c|}{$S^1$} & \multicolumn{3}{c|}{$S^0$} & \multicolumn{3}{c}{$S^1$} \\
& 0-7 & 8-12 & all & 0-7 & 8-12 & all & 0-9 & 10-12 & all & 0-9 & 10-12 & all & 0-11 & 12 & all & 0-11 & 12 & all \\
\hline
BT & 50.12 & - & - & 38.24 & - & - & 47.99 & - & - & 40.97 & - & - & 44.29 & - & - & 44.36 & - & - \\
F\&A & 45.12 & 10.25 & 31.71 & 39.13 & 46.95 & 42.14 & 42.04 & 3.52 & 33.15 & 42.86 & 42.03 & 42.67 & 45.61 & 1.86 & 42.24 & 44.46 & 0.00 & 41.04 \\
FT & 33.96 & 31.83 & 33.14 & 12.34 & 54.13 & 28.41 & 27.80 & 28.01 & 27.85 & 20.79 & 50.55 & 27.66 & 28.18 & 29.98 & 28.32 & 24.06 & 22.33 & 23.93 \\
EWC~\cite{kirkpatrick2017overcoming} & 38.13 & 34.78 & 36.84 & 23.98 & 53.28 & 35.25 & 36.45 & 25.82 & 34.00 & 24.47 & 56.51 & 31.86 & 32.76 & 18.25 & 31.64 & 19.11 & 19.75 & 19.16 \\
LwF~\cite{li2017learning} & 44.51 & 36.14 & 41.29 & 32.82 & 53.47 & 40.76 & 44.68 & 37.69 & 43.07 & 38.29 & 53.33 & 41.76 & 40.38 & \textbf{36.49} & 40.08 & 30.80 & 18.22 & 29.83 \\
GFT+UPG~\cite{yang2023geometry} & 47.36 & 37.84 & 43.70 & 35.43 & \textbf{55.63} & 43.20 & 45.44 & \textbf{41.37} & 44.50 & 37.35 & \textbf{56.97} & 41.88 & 42.79 & 36.38 & 42.30 & 36.00 & \textbf{21.13} & 34.86 \\
\hline
Ours & \textbf{49.12} & \textbf{40.18} & \textbf{45.68} & \textbf{40.07} & 54.05 & \textbf{45.45} & \textbf{46.92} & 37.00 & \textbf{44.63} & \textbf{41.70} & 56.86 & \textbf{45.20} & \textbf{45.74} & 36.36 & \textbf{45.02} & \textbf{40.47} & 20.27 & \textbf{38.92} \\
JT & 51.03 & 37.31 & 45.75 & 37.87 & 59.42 & 46.16 & 47.42 & 40.17 & 45.75 & 42.48 & 58.42 & 46.16 & 46.50 & 36.73 & 45.75 & 46.62 & 40.63 & 46.16 \\
\hline
\end{tabular*}
\end{minipage}
}
\end{table*}

\begin{table*}[t]
\centering 
\renewcommand{\arraystretch}{1.2}
\setlength{\tabcolsep}{3pt}  
\footnotesize 
\caption{The experimental results of 3D point cloud segmentation CIL methods and other necessary comparison methods under split $S^0$ and $S^1$ on the ScanNet dataset, with the best results for CIL methods in this setting highlighted in bold. All results are presented in mIoU (\%).}
\resizebox{0.8\linewidth}{!}{
\begin{minipage}{\linewidth}  
\label{Tab: 2}
\begin{tabular*}{\linewidth}{@{\extracolsep{\fill}}c|ccc|ccc|ccc|ccc|ccc|ccc@{}}
\hline
\multirow{3}{*}{Methods} & \multicolumn{6}{c|}{$C_{novel}=5$} & \multicolumn{6}{c|}{$C_{novel}=3$} & \multicolumn{6}{c}{$C_{novel}=1$} \\
\cline{2-19}
& \multicolumn{3}{c|}{$S^0$} & \multicolumn{3}{c|}{$S^1$} & \multicolumn{3}{c|}{$S^0$} & \multicolumn{3}{c|}{$S^1$} & \multicolumn{3}{c|}{$S^0$} & \multicolumn{3}{c}{$S^1$} \\
& 0-14 & 15-19 & all & 0-14 & 15-19 & all & 0-16 & 17-19 & all & 0-16 & 17-19 & all & 0-18 & 19 & all & 0-18 & 19 & all \\
\hline
BT & 36.23 & - & - & 26.27 & - & - & 33.58 & - & - & 27.61 & - & - & 30.63 & - & - & 30.48 & - & - \\
F\&A & 33.79 & 2.79 & 26.04 & 22.97 & 12.53 & 20.36 & 32.24 & 0.29 & 27.42 & 24.68 & 8.00 & 22.18 & 30.15 & 0.44 & 28.66 & 29.80 & 0.00 & 28.31 \\
FT & 12.18 & 12.74 & 12.32 & 4.18 & 34.47 & 11.75 & 9.01 & 13.41 & 9.67 & 1.65 & 31.05 & 6.06 & 10.52 & 12.18 & 10.63 & 3.80 & 15.59 & 4.39 \\
EWC~\cite{kirkpatrick2017overcoming} & 14.11 & 13.35 & 13.92 & 11.77 & 35.72 & 17.76 & 14.21 & 13.37 & 14.08 & 7.77 & 32.71 & 11.51 & 15.74 & 4.34 & 15.17 & 9.63 & 15.37 & 9.92 \\
LwF~\cite{li2017learning} & 27.62 & 13.43 & 24.07 & 22.24 & \textbf{37.91} & 26.16 & 25.99 & 13.80 & 24.16 & 20.89 & \textbf{45.14} & 24.53 & 22.81 & 13.40 & 22.34 & 21.07 & 15.15 & 20.77 \\
GFT+UPG~\cite{yang2023geometry} & 30.98 & \textbf{14.29} & 26.81 & 23.45 & 34.38 & 26.18 & 26.41 & \textbf{14.82} & 24.67 & 25.51 & 33.71 & 26.74 & 24.65 & 11.60 & 24.00 & 24.43 & \textbf{16.13} & 24.30 \\
\hline
Ours & \textbf{35.35} & 13.98 & \textbf{30.01} & \textbf{28.09} & 37.14 & \textbf{29.45} & \textbf{33.49} & 14.61 & \textbf{30.66} & \textbf{26.91} & 37.03 & \textbf{28.43} & \textbf{28.08} & \textbf{11.99} & \textbf{27.28} & \textbf{27.41} & 15.36 & \textbf{26.81} \\
JT & 36.18 & 16.13 & 31.17 & 28.65 & 41.33 & 31.82 & 33.73 & 16.68 & 31.17 & 29.98 & 42.26 & 31.82 & 31.68 & 21.53 & 31.17 & 31.82 & 31.85 & 31.82 \\
\hline
\end{tabular*}
\end{minipage}
}
\end{table*}

\textcolor{black}{Furthermore, we use adaptive thresholding to determine pseudo-label reliability.} To address the varying uncertainty levels across different class regions and mitigate errors in pseudo-label propagation, particularly in areas where classes overlap and are prone to misclassification, we develop a context-aware adaptive threshold mechanism that accounts for local point distribution characteristics. Our pseudo-label generation follows a hierarchical decision process that selectively integrates knowledge from base and novel models. The decision process prioritizes reliable predictions from the base classifier while handling uncertain regions through neighborhood information.

This hierarchical approach addresses four key scenarios by: (a) trusting high-confidence base classifier predictions when they disagree with initial background classifications; (b) leveraging reliable neighborhood information when the base classifier is uncertain; (c) preserving existing non-background novel class labels; and (d) excluding ambiguous points to maintain training quality. For neighborhood-based label retrieval, we select neighbor points $q$ that satisfy both non-background prediction ($Q_{base}^{q,y} \neq y_{bg}$) and low uncertainty ($\mathcal{U}_n^q \leq \tau$). This selective approach deliberately excludes regions where both target points and their neighbors exhibit high uncertainty, as these typically contain noise that would degrade model performance. The generated pseudo-labels are used to train the novel classifier with the segmentation loss. 

\begin{figure*}[t]
    \centering
    \includegraphics[width=.8\linewidth]{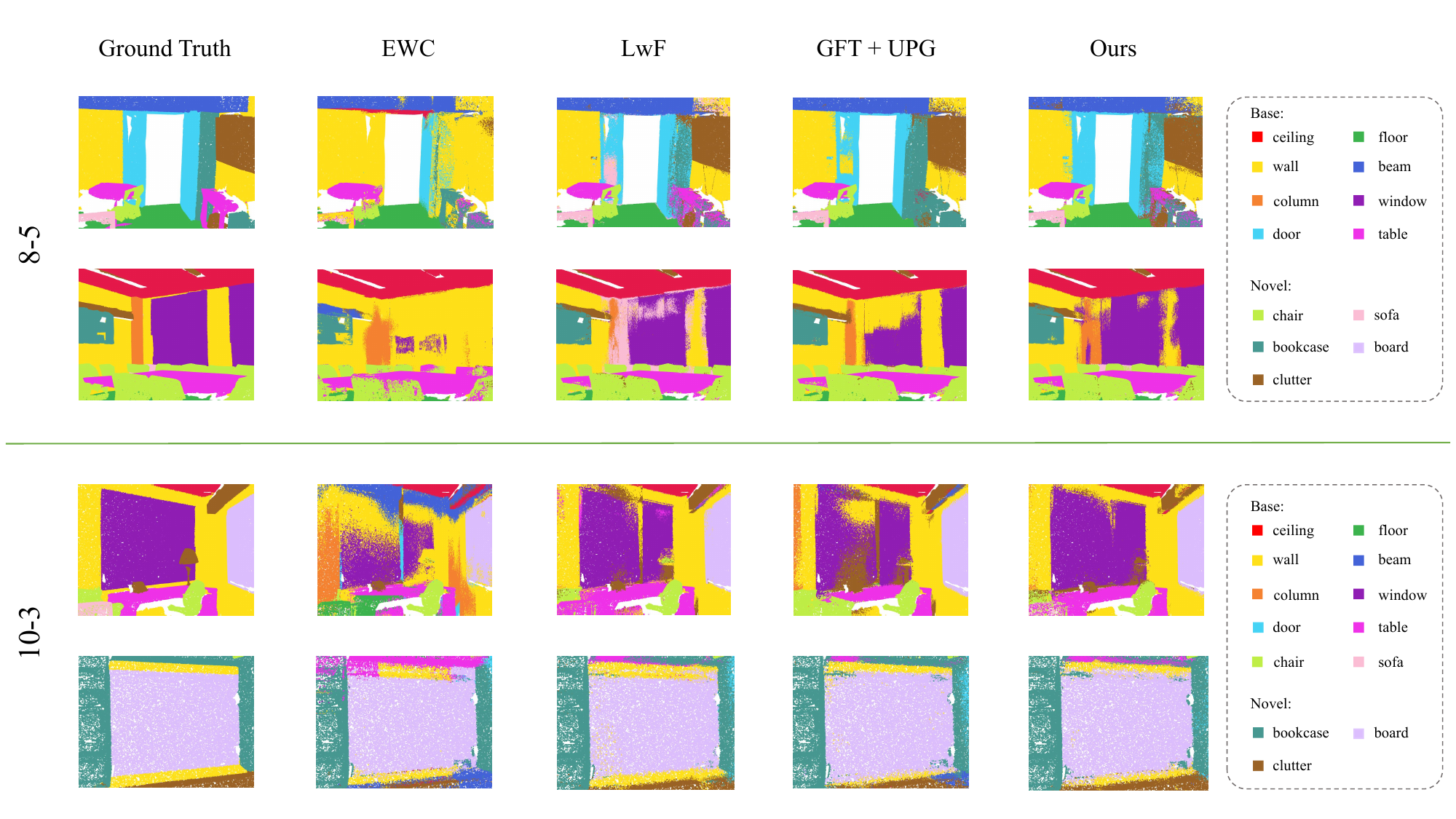}
    \caption{The visualization of the segmentation results from our method, compared with other CIL methods EWC~\cite{kirkpatrick2017overcoming}, LwF~\cite{li2017learning}, and GFT+UPG~\cite{yang2023geometry}, on S3DIS with $C_{\text{novel}}=5$ and $C_{\text{novel}}=3$ for split = 0, shows that our method's segmentation results are closer to the ground truth.}
    \label{Fig: vis}
\end{figure*}

\section{Experiments}
\subsection{Experimental Settings}
\noindent $\textbf{Dataset.}$ In our experiments, we utilize two widely used publicly available datasets: S3DIS~\cite{armeni20163d} and ScanNet~\cite{dai2017scannet}, chosen for their diversity and relevance to the task of point cloud semantic segmentation. S3DIS consists of point clouds from 272 rooms across six different indoor areas. Each point in the dataset includes both XYZ coordinates and RGB color information. The points are annotated with one of 13 predefined semantic classes. Following~\cite{yang2023geometry}, we select Area 5 as the validation set for our experiments, while the remaining areas are used for training. ScanNet is an RGB-D video dataset that includes 1,513 scans taken from 707 indoor scenes. Each point in the dataset is labeled with one of 21 classes, which include 20 semantic classes and an additional class for unannotated areas. We use the 1,210 scans in the dataset for training, while the remaining 312 scans are reserved for validation. For both datasets, we follow the standard training and validation splits and the number of novel classes as in~\cite{yang2023geometry}, ensuring a fair comparison with benchmark methods.

\begin{table*}[t]
\centering
\renewcommand{\arraystretch}{1.2}
\caption{A comparison of the multi-step CIL results between GFT+UPG \cite{yang2023geometry} and our method under the $C_{\text{novel}}=5$ setting on the S3DIS dataset, where each incremental step trains one novel class, with the training epochs and other parameters kept the same as in the single-step CIL setup. The best results are highlighted in bold. All results are presented in mIoU (\%).}
\label{Tab: 4}
\resizebox{.8\textwidth}{!}{
\footnotesize
\begin{tabular}{c|cc|cc|cc|cc|cc}
\hline
\multirow{2}{*}{Class} & \multicolumn{2}{c|}{Step 1} & \multicolumn{2}{c|}{Step 2} & \multicolumn{2}{c|}{Step 3} & \multicolumn{2}{c|}{Step 4} & \multicolumn{2}{c}{Step 5} \\
\cline{2-11}
& GFT+UPG & Ours & GFT+UPG & Ours & GFT+UPG & Ours & GFT+UPG & Ours & GFT+UPG & Ours \\
\hline
0 & 87.85 & 87.96 & 84.48 & 84.29 & 84.72 & 88.63 & 84.94 & 88.30 & 88.21 & 90.19 \\
1 & 95.54 & 95.88 & 96.06 & 95.28 & 96.64 & 96.38 & 96.18 & 96.22 & 96.32 & 97.20 \\
2 & 71.43 & 72.65 & 66.76 & 68.75 & 66.64 & 68.70 & 62.79 & 67.20 & 60.43 & 67.13 \\
3 & 0.09 & 0.30 & 0.27 & 0.76 & 0.28 & 0.23 & 0.38 & 1.50 & 0.60 & 2.61 \\
4 & 6.58 & 0.84 & 7.09 & 0.12 & 5.85 & 2.70 & 10.81 & 3.98 & 7.28 & 3.90 \\
5 & 41.40 & 36.40 & 33.39 & 27.58 & 32.39 & 26.19 & 15.60 & 21.84 & 12.14 & 23.90 \\
6 & 18.72 & 15.67 & 3.05 & 10.51 & 1.00 & 8.70 & 1.33 & 4.92 & 1.31 & 8.10 \\
7 & 62.05 & 61.57 & 48.50 & 56.97 & 47.56 & 54.56 & 32.10 & 51.20 & 33.98 & 54.72 \\
\hline
8 & 31.80 & 30.89 & 5.26 & 20.16 & 3.12 & 20.22 & 4.01 & 20.57 & 3.06 & 20.13 \\
9 & - & - & 3.73 & 3.03 & 3.52 & 2.87 & 4.88 & 3.72 & 3.98 & 3.97 \\
10 & - & - & - & - & 38.11 & 41.13 & 32.50 & 35.27 & 30.46 & 34.30 \\
11 & - & - & - & - & - & - & 8.02 & 8.45 & 8.15 & 8.94 \\
12 & - & - & - & - & - & - & - & - & 35.61 & 36.23 \\
\hline
base & \textbf{48.06} & 46.40 & 42.45 & \textbf{43.04} & 41.87 & \textbf{43.26} & 36.80 & \textbf{41.90} & 36.71 & \textbf{43.47} \\
novel & \textbf{31.80} & 30.89 & 4.65 & \textbf{11.60} & 14.92 & \textbf{21.41} & 12.20 & \textbf{17.00} & 16.25 & \textbf{20.71} \\
all & \textbf{46.25} & 44.68 & 34.89 & \textbf{36.75} & 34.52 & \textbf{37.30} & 28.60 & \textbf{33.60} & 28.84 & \textbf{34.72} \\
\hline
\end{tabular}
}
\end{table*}

\begin{table}[t]
\centering
\renewcommand{\arraystretch}{1.2}
\caption{Performance improvements from different components under the $S^1$, $C_{novel}$=5 and $S^1$, $C_{novel}$=3 settings on S3DIS. FT, PG, and PRO represent Fine-Tuning, ProtoGuard, and Progressive Refinement of Pseudo-Labels, respectively. The best results are highlighted in bold. All results are presented in mIoU (\%).}
\label{Tab: 3}
\resizebox{.9\linewidth}{!}{
\begin{tabular}{ccc|ccc|ccc}
\hline
\multirow{2}{*}{FT} & \multirow{2}{*}{PG} & \multirow{2}{*}{PRO} & \multicolumn{3}{c|}{$S^1$ ($C_{novel}$=5)} & \multicolumn{3}{c}{$S^1$ ($C_{novel}$=3)} \\ 
   &     &     & 0-7  & 8-12 & all  & 0-9 & 10-12 & all  \\ \hline
\checkmark & \xmark & \xmark & 12.34 & 54.13 & 28.41 & 20.79 & 50.55 & 27.66 \\
\checkmark & \checkmark & \xmark & 20.39 & 53.48 & 33.12 & 26.13 & 50.04 & 31.65 \\
\checkmark & \xmark & \checkmark & 34.07 & 54.22 & 41.82 & 35.83 & 57.35 & 40.8 \\ \hline
\checkmark & \checkmark & \checkmark & \textbf{40.07} & \textbf{54.05} & \textbf{45.45} & \textbf{41.70} & \textbf{56.86} & \textbf{45.20} \\ \hline
\end{tabular}
}
\end{table}

\subsection{Performances and Comparisons}
\noindent $\textbf{Comparisons with different methods.}$ 
As shown in Tab~\ref{Tab: 1} and Tab~\ref{Tab: 2}, we provide a detailed comparison of our method with other methods on the two datasets, S3DIS and ScanNet. ``BT" refers to Base Training, ``F\&A" refers to freezing the base model and adding a new classifier output layer during novel class training, and ``FT" refers to randomly initializing the new classifier's last layer and fine-tuning the base model. The above two methods are direct adaptation methods. ``EWC"~\cite{kirkpatrick2017overcoming} and ``LwF"~\cite{li2017learning} are two classic class-incremental learning (CIL) methods, extended here to the 3D point cloud segmentation CIL scenario for comparison. ``GFT+UPG"~\cite{yang2023geometry} is the current state-of-the-art. ``JT" represents joint training, where all classes are trained together and serves as our comparison upper bound. The tables show that the freeze and add method performs excellently on the base classes, even achieving the best performance among all methods in some settings, but performs poorly on the novel classes. This is because it freezes the base class model, preventing it from adapting to the arrival of the novel classes. In contrast, the fine-tuning method frees the corresponding parameters, allowing the model to learn new class knowledge, but this also leads to catastrophic forgetting of the base classes. For CIL methods, a balance between performance on base and novel classes is achieved through various techniques, and none of these methods involve data replay or rehearsal. 

Comparing all CIL methods, our method achieves the best performance on base class testing and overall performance across all scenarios, closely matching the results of joint training, with only a slight performance gap on novel classes in some settings compared to GFT+UPG. 

\textcolor{black}{Fig.~\ref{Fig: vis} presents a visual comparison of segmentation results under the $S^0$ setting with $C_{\text{novel}}=5$ and $C_{\text{novel}}=3$ scenarios. In both scenarios, our method (Ours) outperforms other methods in accurately segmenting both base and novel classes.
In the $C_{\text{novel}}=5$ scenario, Ours effectively distinguishes between base classes like ceiling, wall, and table while maintaining high precision in segmenting novel classes such as chair and sofa. Compared to other methods like EWC, LwF, and GFT+UPG, our method shows significantly less confusion in these classes, particularly in more challenging base class regions like beam and column.
In the $C_{\text{novel}}=3$ scenario, our method also demonstrates superior segmentation, particularly in novel classes such as bookcase and clutter, where other methods tend to misclassify.} 

\noindent $\textbf{Performance in different class orders.}$ To explore the impact of the order of class introduction on incremental segmentation performance, we conducted extensive experiments on two settings, $S^0$ and $S^1$, as shown in Tab~\ref{Tab: 1} and Tab~\ref{Tab: 2}. As can be seen, in the $S^0$ setting, the performance of all methods is better than $S^1$, except in the scenario of  $C_{novel}$=3 on ScanNet. This suggests that the order of class introduction, the semantic correlation of class distributions, and the distribution of feature spaces all influence the performance of incremental segmentation. Taking the S3DIS dataset as an example, classes like table and chair are adjacent in the original order, and the feature distribution of related classes is more concentrated and continuous, which helps the model learn better feature representations. Randomly shuffling the class introduction order would increase the task's difficulty. However, it can still be observed that our method achieves the best performance across different orders.

\subsection{Ablation Studies}
\noindent $\textbf{Effect of ProtoGuard and PROPEL.}$ \textcolor{black}{To investigate the impact of each module on the performance of class-incremental learning, we conducted a comprehensive ablation study for both modules under the $S^1$ setting with $C_{\text{novel}}=5$ and $C_{\text{novel}}=3$ scenarios on the S3DIS dataset, as shown in Table~\ref{Tab: 3}.
Even more remarkably, when PROPEL is integrated with fine-tuning, base class performance jumps dramatically by 21.73 mIoU points to 34.07\%. 
The combination of both modules (FT+PG+PRO) yields the best results, with base class performance reaching 40.07 mIoU (+27.73 mIoU points over FT alone) and overall performance of 45.45 mIoU (+17.04 mIoU points). This effect demonstrates that ProtoGuard provides more discriminative features that improve PROPEL's uncertainty estimation and pseudo-label generation quality.}

\noindent $\textbf{Multi-step setting.}$ \textcolor{black}{To examine a more challenging scenario with increased incremental steps, we performed multi-step CIL experiments in the S3DIS dataset with the $S^0$ and $C_{\text{novel}}=5$ scenario following~\cite{yang2023geometry}. As the number of incremental steps increases, the task becomes more difficult due to the longer training duration, intensifying the challenges of mitigating catastrophic forgetting.} From Tab~\ref{Tab: 4}, we can observe that GFT+UPG shows a significant performance drop in classes 6, 7, and 8 as the number of incremental steps increases, whereas our method effectively controls catastrophic forgetting in these classes. These classes are either long tail or located in point clouds' overlapping areas. When they belong to both new and old classes, adding new classes can easily confuse the previously learned knowledge of the old classes. Our approach mitigates this issue by maintaining continuously updated prototypes for each class, effectively preventing this situation.

\section{Conclusions}
In this paper, we propose a two-stage method for 3D point cloud segmentation. In the base-class training phase, we maintain geometric and semantic prototypes for each class and combine them with edge features to generate more discriminative features. In the novel-class training phase, we leverage the feature extractor and classifier trained in the previous stage and provide reliable guidance for pseudo-label generation and propagation by incorporating information such as density distribution. Extensive experiments demonstrate that our method effectively maintains segmentation performance on new classes while avoiding catastrophic forgetting of base classes. 

{\small
\bibliographystyle{IEEEtran}
\bibliography{ieee}
}
\end{document}